\DocumentMetadata{
      pdfversion=2.0,pdfstandard=ua-2,
      testphase={phase-III,firstaid,math,title}}
\documentclass[sigconf, nonacm]{acmart-tagged}

\setcopyright{none}
\renewcommand\footnotetextcopyrightpermission[1]{} % Removes the copyright footnote

\AtBeginDocument{%
  }

\acmYear{2026}
\copyrightyear{2026}
\acmConference[HRI Companion '26]{Companion Proceedings of the 21st ACM/IEEE International Conference on Human-Robot Interaction}{March 16--19, 2026}{Edinburgh, Scotland, UK}
\received{2025-12-08}
\received[accepted]{2026-01-12}

\usepackage{multirow}
\usepackage[most]{tcolorbox}
\usepackage[utf8]{inputenc} 
\usepackage{balance} 

\begin{document}

\title{Human-Aware Robot Behaviour in Self-Driving Labs}

\author{Satheeshkumar Veeramani}
\authornote{These authors contributed equally to this work.}
\affiliation{
  \department{Department of Chemistry}
  \institution{University of Liverpool}
  \country{United Kingdom}}
\email {satheeshkumar.veeramani@liverpool.ac.uk}

\author{Anna Kisil}
\authornotemark[1]
\affiliation{
  \department{School of Computer Science and Informatics}
  \institution{University of Liverpool}
  \country{United Kingdom}}
\email {anna.kisil@liverpool.ac.uk}

\author{Abigail Bentley}
\affiliation{
  \department{Department of Chemistry}
  \institution{University of Liverpool}
  \country{United Kingdom}}
\email {a.r.bentley@liverpool.ac.uk}

\author{Hatem Fakhruldeen}
\affiliation{
  \department{Department of Chemistry}
  \institution{University of Liverpool}
  \country{United Kingdom}}
\email {h.fakhruldeen@liverpool.ac.uk}

\author{Gabriella Pizzuto}
\affiliation{
\department{School of Computer Science and Informatics \\ Department of Chemistry}
  \institution{University of Liverpool}
  \country{United Kingdom}}
\email {gabriella.pizzuto@liverpool.ac.uk}

\author{Andrew I. Cooper}
\affiliation{
\department{Department of Chemistry}
  \institution{University of Liverpool}
  \country{United Kingdom}}
\email {aicooper@liverpool.ac.uk}

\renewcommand{\shortauthors}{S.Veeramani, A.Kisil, A. Bentley, H.Fakhruldeen, G.Pizzuto, A.I.Cooper}

\begin{abstract}
Self-driving laboratories (SDLs) are rapidly transforming research in chemistry and materials science to accelerate new discoveries.
Mobile robot chemists (MRCs) play a pivotal role by autonomously navigating the lab to transport samples, effectively connecting synthesis, analysis, and characterisation equipment.
The instruments within an SDL are typically designed or retrofitted to be accessed by both human and robotic chemists, ensuring operational flexibility and integration between manual and automated workflows.
In many scenarios, human and robotic chemists may need to use the same equipment simultaneously. 
Currently, MRCs rely on simple LiDAR-based obstruction detection, which forces the robot to passively wait if a human is present. 
This lack of situational awareness leads to unnecessary delays and inefficient coordination in time-critical automated workflows in human-robot shared labs.
To address this, we present an initial study of an embodied, AI-driven perception method that facilitates proactive human-robot interaction in shared-access scenarios.
Our method features a hierarchical human intention prediction model that allows the robot to distinguish between preparatory actions (waiting) and transient interactions (accessing the instrument).
Our results demonstrate that the proposed approach enhances efficiency by enabling proactive human–robot interaction, streamlining coordination, and potentially increasing the efficiency of autonomous scientific labs.
\end{abstract}

% \begin{CCSXML}
% <ccs2012>
%    <concept>
%        <concept_id>10003120.10003121.10003122.10011749</concept_id>
%        <concept_desc>Human-centered computing~Laboratory experiments</concept_desc>
%        <concept_significance>500</concept_significance>
%        </concept>
%    <concept>
%        <concept_id>10010147.10010178.10010199.10010204</concept_id>
%        <concept_desc>Computing methodologies~Robotic planning</concept_desc>
%        <concept_significance>500</concept_significance>
%        </concept>
%    <concept>
%        <concept_id>10010147.10010178.10010199.10010201</concept_id>
%        <concept_desc>Computing methodologies~Planning under uncertainty</concept_desc>
%        <concept_significance>500</concept_significance>
%        </concept>
%    <concept>
%        <concept_id>10010147.10010178.10010224.10010225.10011295</concept_id>
%        <concept_desc>Computing methodologies~Scene anomaly detection</concept_desc>
%        <concept_significance>500</concept_significance>
%        </concept>
% </ccs2012>
% \end{CCSXML}

% \ccsdesc[500]{Human-centered computing~Laboratory experiments}
% \ccsdesc[500]{Computing methodologies~Robotic planning}
% \ccsdesc[500]{Computing methodologies~Planning under uncertainty}
% \ccsdesc[500]{Computing methodologies~Scene anomaly detection}

% \keywords{Human-Robot Interaction, Intelligent Lab Automation}

\maketitle

\begin{figure}[h]
    \centering
    \includegraphics[width=1.2\columnwidth, alt={A cartoon illustration depicting a collaboration between a robotic chemist and a human chemist in a laboratory. On the left, a mobile KUKA robot with an articulated arm carries a tray of small vials. On the right, a human chemist wearing a white lab coat, safety goggles, and gloves stands holding a smartphone. In the background, there is a large fume hood containing chemical flasks. A dialogue box indicates a conversation between the two: the Robotic Chemist asks, "You seem to be using the fumehood. Shall I wait until you're done?" and the Human Chemist replies, "Yes, please wait a moment."}]{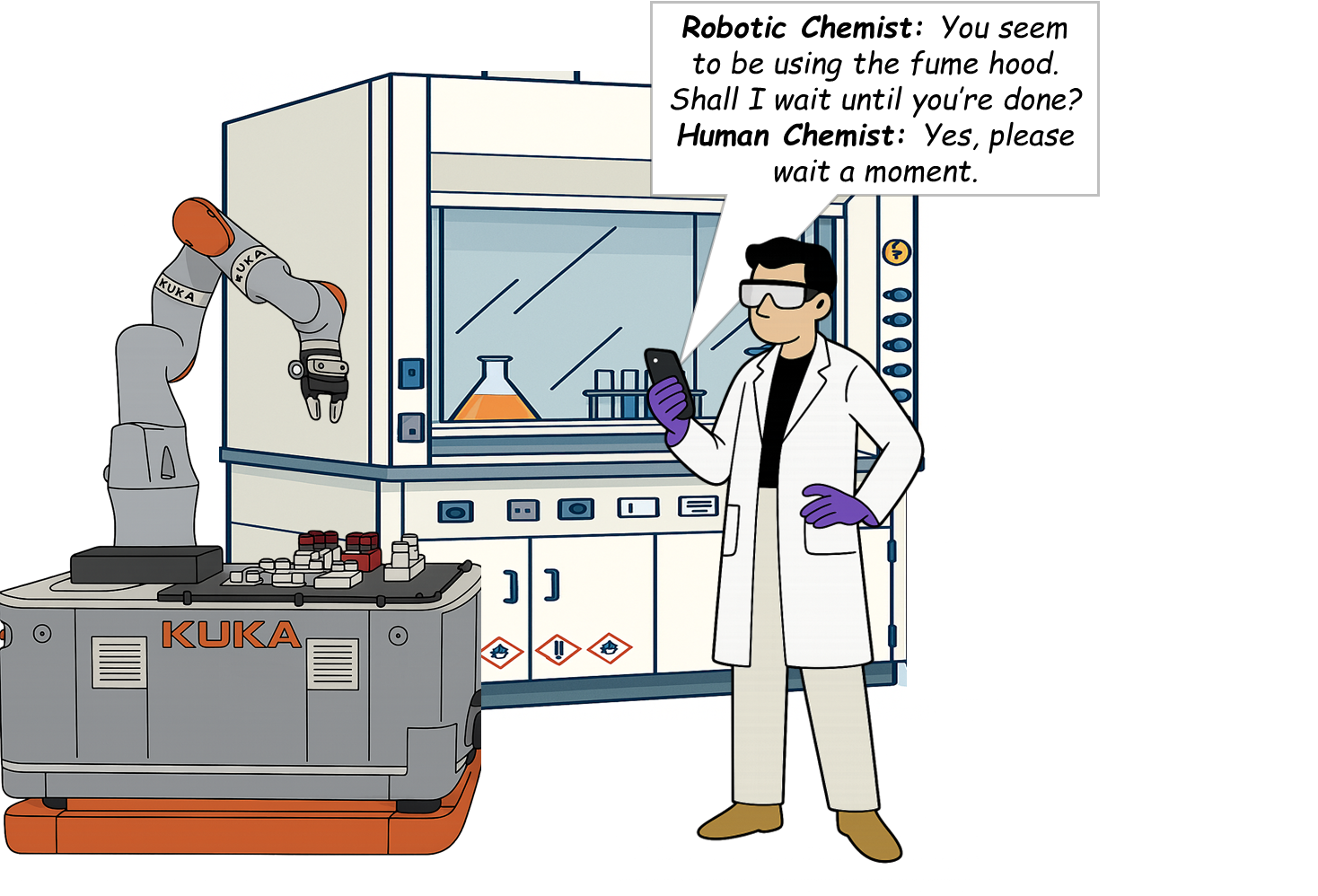}
    \caption{Shared workspace interaction between a human chemist and a collaborative mobile robot near a standard laboratory fumehood.}
    \label{fig:teaser}
\end{figure} 

\section{Introduction}
In SDLs, robotic chemists and human researchers play key, albeit potentially different, roles in experimental workflows~\cite{orouji2025autonomous}.
Herein, MRCs are now capable of running autonomous workflows, mainly by transferring samples between various laboratory equipment for tasks such as solid/liquid handling, synthesis (\textit{e.g.}, Chemspeed platforms), and analysis (\textit{e.g.}, liquid chromatography–mass spectrometry)~\cite{burger2020mobile, brass2025mobile, dai2024autonomous}. 
Human chemists, who work in the same environment as robotic chemists, primarily design experiments, prepare and set up the chemicals for the workflow, and oversee the execution of the experiment to mitigate failures and to ensure safety and reliability. To date, MRCs typically receive tasks from a centralised orchestration software that coordinates workflows using communication frameworks such as ROS~\cite{fakhruldeen2022archemist, EOS, fernando2025robotic}, relying on a local task execution framework within the robot controller. Despite advancements in SDLs, current MRCs operate without awareness of human intent. This setup fundamentally precludes dynamic interaction and shared decision-making with human chemists. 

This open gap creates a critical opportunity for the human-robot interaction (HRI) community to contribute by designing and implementing systems where MRCs can perceive, predict, and respond to human intent, thereby enabling more fluent interaction, safer task sharing, and improved overall workflow efficiency.
Existing social navigation frameworks focus primarily on replanning the trajectory to avoid humans and obstacles, which cannot be applied to SDLs due to tighter space constraints.
It becomes ever more critical when both human and robotic chemists need to access shared laboratory instruments or move within overlapping work zones~\cite{prevent}. 
Instead, robot chemists should use perception to proactively interact usefully with human collaborators~\cite{olivianav, LookFurther}. 
Without this ability to anticipate human activity, communicate intent, or adjust their behaviour accordingly, MRCs risk interrupting ongoing experiments or compromising safety~\cite{lagomarsino2025intuitive}.

To address this gap, we propose an embodied, AI-driven perception, reasoning and interaction method that enables robots to detect human presence, infer human engagement, and initiate proactive communication to reduce unnecessary waiting time and improve workflow fluency in shared laboratory scenarios. 
The method employs a hierarchical two-stage human detection model that can be coupled with a text-based user interface, supporting adaptive, context-aware navigation and facilitating effective human-robot interaction.

\section{Related Work}
In autonomous scientific labs, decision-level fusion of information from modalities has been shown to be effective for various robotic tasks, including task execution and monitoring ~\cite{fakhruldeen2025multimodal, DARVISH2025101897}, anomaly detection ~\cite{prevent}, self-correction ~\cite{li2024self}, and analytical evaluation ~\cite{pizzuto2022solis, SHIRI2021102176}. 
In these approaches, individual models provide separate predictions, and the final decision is derived through either majority voting or weighted aggregation of modalities. 
However, with recent advances in artificial intelligence, large language models (LLMs) have demonstrated the capability to perform data fusion at intermediate levels through multimodal embeddings~\cite{yin2024survey}. 
This approach enhances cross-modal reasoning and context-aware learning while mitigating decision conflicts that typically arise in isolated models~\cite{cooper2025lira}. Such context-aware intelligence becomes particularly important when robots need to understand human behaviour in shared environments. For instance, in social navigation problems, robots rely primarily on real-time person-tracking modules to detect and localise humans with respect to the robot’s position, after which the robot replans its path accordingly~\cite{LookFurther}. 
However, previous works~\cite{olivianav, vlmsocialnav} have shown that contextual information about the scene provides essential social reasoning cues that guide robot actions. 
This is particularly important in work environments such as laboratories, where both humans and robots simultaneously operate within the same shared workspace. 
Our work extends these ideas to safety-critical laboratory environments by enabling mobile robotic chemists to interpret contextual cues and human intentions, supporting proactive, socially-aware navigation.
\section{Method}
\begin{figure*}[t]
\centering
    \includegraphics[width=0.8\linewidth, alt={A technical diagram illustrating an AI system architecture divided into two main sections: "Multimodal perception" and "Vision-language reasoning." On the left, RGB images and depth maps are processed by a detection module using YOLO and ROS. This data flows to the right section, where it is combined with text prompts describing a lab scene. These inputs pass through an image encoder and tokeniser into a "LLaVA-1.5-7b" model. The final output provides status indicators (e.g., "Busy: Yes") and a natural language response: "You seem to be using the fume hood. Shall I wait until you're done?"}]{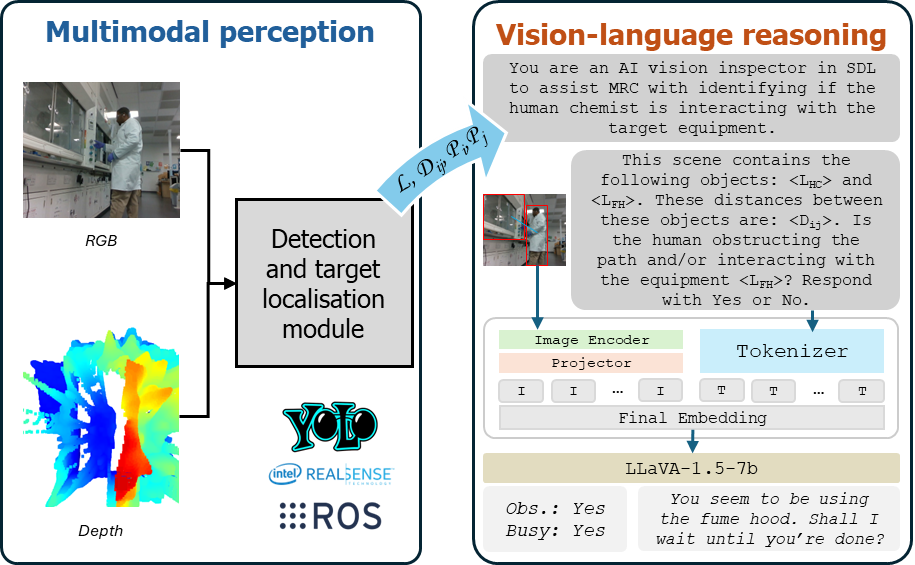}
    \caption{Proposed method to predict the intention of human chemists in a shared workspace within an SDL. $L$ denotes the set of detected labels. $P_i$ and $P_j$ represent the 3D position vectors of objects $i$ and $j$, respectively. $D_{ij}$ is the Euclidean distance between objects $i$ and $j$.}
    \label{fig:SA}
\end{figure*}

This section presents the proposed methodology (Figure~\ref{fig:SA}) designed for MRCs to detect the presence of human chemists within their workspace, understand the context of the scientists' activity, and take appropriate action. 
This section also details the proposed architecture, which consists of two stages: perception and reasoning. 
In the first stage, multiple sensory modalities continuously monitor the workspace to detect the presence and location of human chemists and other equipment. 
Then, once human presence is detected, the second stage focuses on the perceived and estimated data, which is forwarded to a vision–language model (VLM) that interprets the context of the interaction and generates an appropriate message to communicate with the human chemist.

% While this work focuses on a laboratory setting, the proposed method is not specific to this environment and is transferable to other shared workspace HRI contexts, given an appropriate domain-specific dataset.

\subsection{Multimodal Perception Module}

The proposed method is initiated when the robot receives a task request from the central orchestration system. 
As the robot begins navigating toward its destination, the vision and depth modalities, along with the built-in LiDAR module, simultaneously perceive and analyse its surrounding environment.
An object detection system is employed to detect and classify objects within an image, generating bounding boxes that provide the 2D coordinates of the detected objects. 
The process operates on frames obtained from a video stream captured by a camera mounted on the robot base. 
The stereo depth-sensing module is subsequently used to estimate the third (depth) dimension, enabling 3D localisation of the detected objects. 
Several object classes are defined within the model to identify and filter relevant entities that contribute to understanding the scene context. 
The defined classes are human chemists, standard laboratory instruments and fumehoods.
After obtaining the labels of the detected classes and their corresponding 3D coordinates with respect to the robot’s reference frame, the distances between them (\texttt{$D_{ij}$}) are calculated using equation \ref{DistanceEqn} to assess their spatial proximity and infer contextual relationships between all detected object pairs.

\begin{equation}
D_{ij} = \left\| \mathbf{p}_i - \mathbf{p}_j \right\|_2 ,
\qquad
\forall\, i,j \in \{1,2,\dots,N\},\, i \ne j.
\label{DistanceEqn}
\end{equation}

\noindent
where $\mathbf{p}_i = [x_i,\, y_i,\, z_i]^\top$ represents the 3D position vector of the $i$\textit-th detected object.

\subsection{Vision-Language Reasoning and Interaction}
The VLM module shown in Figure~\ref{fig:SA} illustrates the standardised prompt used to obtain responses from the VLM. 
The model is provided with multi-modal inputs, including the original image, class labels and the rules for handling distance information calculated in Equation~\ref{DistanceEqn}. 
$D_{ij}$ is incorporated into the prompts automatically to enable spatial reasoning within the VLM. 
For example, in the following prompt, \texttt{$L_{Hc}$} denotes the human chemist and \texttt{$L_{FH}$} denotes the fumehood. 
This prompt is used to predict whether the human chemist is obstructing the robot and/or interacting with the fumehood (\textit{i.e.,} the chemist is currently busy and should not be disturbed).

\begin{tcolorbox}[colback=gray!10, colframe=gray!40, arc=2mm, boxrule=0.3pt]
\texttt{This scene contains the following objects: <L$_{HC}$> and <L$_{FH}$>. 
The distances between these objects are: <D$_{ij}$>. Is the human obstructing the path and/or interacting with the equipment (<L$_{FH}$>)? Respond with Yes or No.}\end{tcolorbox}

Initially, the model returns two binary responses: Yes or No, corresponding to both obstruction and interaction. 
Based on these predictions, it further evaluates whether the robot should approach the human chemist and generates the appropriate interaction message.

\begin{tcolorbox}[colback=gray!10, colframe=gray!40, arc=2mm, boxrule=0.3pt]
\texttt{\textbf{Obstruction}: Yes; \textbf{Interaction}: Yes; \textbf{Message}: You seem to be using the fumehood. Shall I wait until you are done?}
\end{tcolorbox}
\section{Experimental Evaluation}
This section presents the preliminary results of the proposed method. 
The initial study involving data collection and testing was conducted in an autonomous chemistry laboratory ~\cite{burger2020mobile, dai2024autonomous}, where MRCs and human chemists operate within a shared workspace. 
This environment provides a realistic and dynamically changing setting to assess the performance and robustness of our human-aware robot behaviour framework. 
We identified three interaction scenarios in which the potential for workspace overlap is high, and evaluated the robustness of the proposed two-stage multi-modal model.

\subsection{Setup and Data Collection}

The experimental setup consists of a KUKA KMR-iiwa (a mobile robotic manipulator) equipped with an Intel RealSense D435i RGB-D camera, mounted on the mobile base. 
An onboard computer with a 64GB RAM is also mounted on the deck to run the proposed method online. 
A machine equipped with 13th Gen Intel\textsuperscript{\textregistered} Core\textsuperscript{TM} i7-13700KF  CPU and an NVIDIA GeForce RTX 4090 GPU was used to fine-tune the models. 
A total of 3270 sets of data (images and depth data) were collected across three interaction scenarios, where (1) a human chemist using equipment that the robot needs to access, (2) a human chemist obstructing the robot’s path, and (3) multiple humans present in front of the robot, doing various lab activities. 
The data collection process was fully automated and executed multiple times across the three interaction types. 
For this initial study, the authors of this paper served as participants for the data collection.

The model used in the vision-language module is the LLaVA-1.5-7b, a multimodal model that has demonstrated strong performance in tasks involving visual understanding and human intent inference~\cite{llava}. 
First, the collected images were categorised into three classes based on two binary variables: obstruction and interaction. Images represented situations where the human was (1) obstructing the path and interacting with equipment, (2) neither obstructing nor interacting, or (3) obstructing but not interacting. 
The fourth possible class (interacting without obstructing) was excluded as it is physically impossible for a human to interact with the equipment without obstructing the robot's path. 
Following image collection and categorisation, a labelled dataset was created, containing expected model outputs to be used for model fine-tuning. 
The binary interaction and obstruction labels were added manually, while the natural language explanations were generated using GPT-4~\cite{gpt4}, prompted to provide concise descriptions of the images. 
The final dataset was split into 5 subsets, and model training was run 5 times, where each subset was used once for testing and the remaining 4 for training. 
The final model test accuracies presented in Table~\ref{tab:modality_accuracy} are the rounded mean and variance of these 5 runs.

\subsection{Model Testing and Evaluation}
% \cite{yolov8_ultralytics}
We evaluate three model configurations: (1) the base model, (2) the fine-tuned model, and (3) the fine-tuned model provided with additional data (i.e. YOLO-derived object labels ($L$), distance measurements ($D_{ij}$) between the human and the equipment as well as between the human and the robot (when a human is present), and text-based reasoning rules that define distance thresholds indicating when the human is interacting with the equipment or obstructing the robot’s path).
The test accuracy is measured by comparing the model's predicted obstruction and interaction labels against ground truth values. 
A prediction is considered correct only when both labels match. As shown in Table~\ref{tab:modality_accuracy}, fine-tuning alone improves performance over the baseline by 59\%, 74\%, and 47\% across the three scenarios, respectively. 

\begin{table}[h]
\centering
\caption{Model Performance}
\label{tab:modality_accuracy}
\begin{tabular}{c|c|c|c}
\hline
\textbf{Scenario} & \textbf{Modality} & \textbf{Model} & \textbf{Test Accuracy (\%)} \\
\hline
\multirow{3}{*}{S1} 
 & Vision & LLaVA-1.5-7b & 29±3 \\
 & Vision & LLaVA-1.5-7b* & 88±4 \\
 & Vision+Depth & LLaVA-1.5-7b* & 70±8 \\
\hline
\multirow{3}{*}{S2} 
 & Vision & LLaVA-1.5-7b & 20±2 \\
 & Vision & LLaVA-1.5-7b* & 94±2 \\
 & Vision+Depth & LLaVA-1.5-7b* & 94±2 \\
\hline
\multirow{3}{*}{S3} 
 & Vision & LLaVA-1.5-7b & 43±2 \\
 & Vision & LLaVA-1.5-7b* & 90±1 \\
 & Vision+Depth & LLaVA-1.5-7b* & 82±2 \\
\hline
\end{tabular}

\vspace{2pt}
\footnotesize
\textit{Notes:} 
S1 – Human obstructing MRC's access to the equipment  (destination); 
S2 – Human obstruction on the path (near fumehood); 
S3 – Multi-human obstructions;\newline
* denotes models fine-tuned with the collected dataset; 
\end{table}

Most failure cases are caused by a lack of sufficient contextual information needed to make accurate inferences in ambiguous situations. 
As a result, two main problems emerge in all three scenarios. 
First, the model occasionally struggles to correctly identify the human's body posture and orientation, leading to incorrect interaction labels. 
Second, the model assumes the robot's goal location to be directly ahead (\textit{i.e.}, in the centre of the image), failing to differentiate between the target equipment and the goal location. 
Consequently, it generates false-positive obstruction labels when equipment is off-centre and false-negative obstruction labels when the human obstructs the path by standing near the target equipment, rather than directly in front of it.
The latter problem could be partially resolved by providing depth-based distance measurements between the chemist, equipment, and robot, allowing the model to assess obstruction and interaction status by comparing observed distances against predefined threshold-based distance rules. 
However, Table~\ref{tab:modality_accuracy} shows that the incorporation of these additional measurements decreases accuracy by 18\% and 8\%, for scenarios 1 and 3, respectively. 

Despite the overall decrease in accuracy, the failure scenarios differ between the two experiments. 
In cases where the observed distance measurement clearly indicates obstruction, the model correctly identifies this in experiment 3 (unlike the false negative produced in the identical case during experiment 2), indicating that the distance information is being used by the model. 
However, incorporating additional information into the prompt increases its complexity, frequently confusing or misleading the model, rather than improving its reasoning across all cases. 
In these instances, the model either fails to correctly interpret the distance measurements or becomes over-reliant on the distance-based rules, essentially `forgetting' any previous reasoning that produced correct results in experiment 2. 
Our results show that although additional distance information can resolve ambiguity in specific cases, the method of simply incorporating this information into the prompt lacks sufficient reliability. 
Therefore, future work will concentrate on addressing this limitation through a retrieval-augmented generation (RAG) approach, whereby additional information (including distance measurements and laboratory topology data) will be supplied to the model in a more reliable way.

% This is likely due to a lack of distinction between the target equipment and the goal location and an increase in prompt complexity, which confuses the model rather than improving its reasoning, and ultimately causes the model to make further false negatives. In practice, we can overcome this either by performing adversarial training on the image encoder or by reducing prompt complexity. One approach to address this limitation in future work is to provide the model with the topology information of the lab via retrieval-augmented generation (RAG).

 % \textcolor{red}{ drop in performance likely stems from the increase in prompt complexity, which confuses the model rather than improving its reasoning, as well as the difficulty in defining accurate universal distance threshold values.}

 % The additional information provided in the third experiment consists of: 
% \begin{itemize}
%     \item YOLO-derived object labels;
%     \item distance measurements ($D_{ij}$) between the human and equipment, and between the human and the robot (when a human is present);
%     \item text-based reasoning rules containing distance thresholds that indicate when the human is interacting with the equipment or obstructing the robot's path.
% \end{itemize}

\section{Conclusion}
The integration of VLMs into MRCs addresses several long-standing challenges in HRI, including perception, reasoning, and communication. 
This paper presents a two-stage, AI-driven, perception-based framework that detects the presence and interprets the intentions of human chemists during robot navigation tasks in an autonomous chemistry laboratory environment. 
The evaluation results demonstrate that incorporating additional distance information and rules would help the method to infer the intentions of human scientists. 
However, our results demonstrated that the model requires the real-time topology information for generalising this approach across broader laboratory scenarios. 
This methodology is expected to improve operational efficiency by minimising idle time, enabling robotic chemists to dynamically reallocate their attention to other productive tasks.
The scope of this work is limited to the design of a two-stage model, scenario identification, data collection, and model validation. 
Future work will focus on implementation, generalisation of the proposed approach to a wider range of laboratory scenarios through adversarial training and RAG, experimental evaluation, and comprehensive user studies. 
These efforts will help assess the model’s scalability, robustness, and impact on workflow efficiency in real-world autonomous laboratory environments.

\begin{acks}
This work was supported by the Leverhulme Trust through the Leverhulme Research Centre for Functional Materials Design, the Royal Academy of Engineering under the Research Fellowship Scheme and EPSRC through the AI for chemistry: AIchemy hub (EP/Y028775/1). A.I.C. thanks the Royal Society for a Research Professorship (RSRP\textbackslash S2\textbackslash 232003).
The authors acknowledge the use of OpenAI’s ChatGPT (DALL-E) for generating some of the illustrative content and visual materials presented in this paper.
\end{acks}
\balance
\bibliographystyle{ieeetr}
\bibliography{bibliography}
\end{document}